\documentclass[conference]{IEEEtran}
\IEEEoverridecommandlockouts
\IEEEoverridecommandlockouts
\usepackage{graphicx}
\usepackage{epsfig,epstopdf}
\usepackage{cite}
\usepackage{multirow}
\usepackage{times}
\renewcommand*\ttdefault{txtt}
\usepackage{soul}
\usepackage{url}
\usepackage[hidelinks]{hyperref}

\usepackage[small]{caption}
\usepackage{subcaption}
\usepackage{amsmath}
\usepackage{booktabs}
\usepackage{algorithm}
\usepackage{algorithmic}
\usepackage{enumerate}
\usepackage{amssymb,amsmath}
\usepackage{color}
\usepackage{comment}
\usepackage{enumerate}
\def\BibTeX{{\rm B\kern-.05em{\sc i\kern-.025em b}\kern-.08em
    T\kern-.1667em\lower.7ex\hbox{E}\kern-.125emX}}
\begin{document}

\title{Deep Incomplete Multi-View Multiple Clusterings\thanks{This paper was accepted by icdm 2020.}}
\author{Shaowei Wei$^{1,2}$, Jun Wang$^{1,2,3,*}$\thanks{$^*$Corresponding author: kingjun@sdu.edu.cn (Jun Wang). This work is supported by NSFC (No. 62072380, 61872300 and 62031003).},  Guoxian Yu$^{1,2,3,4}$, Carlotta Domeniconi$^5$,  Xiangliang Zhang$^4$\\
$^1$Joint SDU-NTU Centre for Artificial Intelligence Research, Shandong University, Jinan, China\\
$^2$College of Computer and Information Sciences, Southwest University, Chongqing, China\\
$^3$School of Software, Shandong University, Jinan, China\\
$^4$CEMSE, King Abdullah University of Science and Technology, Thuwal, SA \\
$^5$Department of Computer Science, George Mason University, VA, USA\\
Email: swwei2019@email.swu.edu.cn; \{kingjun,gxyu\}@sdu.edu.cn, carlotta@cs.gmu.edu; xiangliang.zhang@kaust.edu.sa\\
}

\maketitle

\begin{abstract}
Multi-view clustering aims at exploiting information from multiple heterogeneous views to promote clustering. Most previous works search for only one optimal clustering based on the predefined clustering criterion, but devising such a criterion that captures what users need is difficult. Due to the multiplicity of multi-view data, we can have meaningful alternative  clusterings. In addition, the incomplete multi-view data problem is ubiquitous in real world but has not been studied for multiple clusterings.
To address these issues, we introduce a deep incomplete multi-view multiple clusterings (DiMVMC) framework, which achieves the completion of data view and multiple shared representations simultaneously by optimizing multiple groups of decoder deep networks. In addition, it minimizes a redundancy term to simultaneously
control the diversity among these representations and among parameters of different networks. Next, it generates an individual clustering from each of these shared representations. Experiments on benchmark datasets confirm that DiMVMC outperforms the state-of-the-art competitors in generating multiple clusterings with high diversity and quality.
\end{abstract}


\section{Introduction}
With the wide-application of Internet of Things, many collected data are naturally represented with multiple feature views. For instance, an image can be encoded by its color, texture, shape and spatial descriptors. These feature views embody the consistent and complementary information of the same image, which spur extensive research of learning on multi-view data \cite{sun2013mvlsurvey,yang2018multi}. Fusing these feature views can not only form a comprehensive description of the data, but also benefit the learning tasks on them, such as classification \cite{tan2018incomplete}, clustering \cite{kang2019aligned} and metric learning \cite{hu2017sharable}. This work focuses on multi-view clustering (MVC), which aims to excavate complementary and consensus information across multiple views to identify the essential grouping structure with no requirement of labels from these data.

Various attempts have been made to find essential grouping structures of multi-view data. Some algorithms force the clustering results of different views being consistent with each other  via  correlation maximization \cite{chaudhuri2009multi}, co-regularization \cite{kumar2011coreg}, fusing multiple similarity matrices of individual views \cite{liu2016multiple}, {or exploring common and diverse information among views \cite{2015Partially}}. Other approaches assume the clusters of a clustering are embedded in different subspaces and try to explore these subspaces and find clusters therein \cite{parsons2004subspace,luo2018consistent}. More recently, deep learning techniques also have been proposed to extract the high-order correlation and dependency between multi-view data for effective clustering \cite{li2019deep,zhao2017dmf}. Besides, some other efforts study how to work on multi-view data with missing views, i.e., some objects are not available for all the views \cite{li2014partial,xu2019adversarial}.


These mentioned MVC algorithms account for the multiplicity of multi-view data, but focus on generating a single clustering result only. In practice, such multiplicity can also support to group the data in multiple different but meaningful clusterings \cite{fanaee2018multi,yao2019MVMC}. For example, a bunch of facial images represented with heterogeneous views can be separately grouped from the perspective of \emph{identity, sex} and of \emph{emotions}. All these groupings are different but yet meaningful. These reasonable groupings of the data are potentially useful for some purposes, regardless of whether or not it is optimal according to a specific clustering criterion \cite{caruana2006meta}. Alike traditional clustering that focuses on the  quality, \textbf{multiple clusterings} additionally pursue the diversity among alternative clusterings. However, it is a knotty task to balance the diversity and quality of these clusterings \cite{bailey2013alternative}. {Previous works tried to obtain multiple clusterings in independent (or orthogonal) subspaces \cite{cui2007OSC,mautz2018NrKMeans,wang2019MISC}, by eliminating redundancy between the clusterings that are generated successively \cite{bae2006coala,yang2017MNMF}, by executing clustering assignment again for the generated base clusterings\cite{caruana2006meta}, or by simultaneously gaining multiple clusterings and controlling the redundancy. However, they were designed for \textbf{single-view} data only.}


A few efforts have been made toward exploring multiple clusterings on \textbf{multi-view} data. {Multi-view multiple clusterings (MVMC) \cite{yao2019MVMC} mines the individual and shared information of multi-view data by utilizing self-representation learning \cite{luo2018consistent}, and then decomposes the combinations of the individuality feature matrices and commonality feature matrix by semi-nonnegative matrix factorization \cite{ding2010convex} to obtain multiple clusterings. DMClusts\cite{wei2019multi} is another multi-view multiple clusterings algorithm based on deep matrix factorization. It decomposes the multi-view data matrices layer-by-layer to obtain multiple common subspaces and generate corresponding clusterings therein.} These two efforts still ideally assume all the data views are complete. However, this assumption is often violated in practice for some inevitable reasons \cite{li2014partial,tan2018incomplete}, such as the temporary failure of sensors or the human caused errors. As a result, the collected multi-view data are often \textbf{incomplete}. A simple strategy is to remove the samples with missing feature views, but this strategy obviously may throw away too much information, especially when with a high missing rate. Incomplete multi-view clustering (IMC) solutions have been proposed to address this practical issue. Some of them resort to matrix factorization to extract the shared subspace \cite{li2014partial,xu2018partial}, fill the missing information \cite{hu2019IMC}, or use Generate Adversarial Network (GAN) \cite{goodfellow2014gan} to replenish the missing data \cite{wang2018partial,xu2019adversarial}. However, none of existing IMC methods can  generate multiple clusterings with both high quality and diversity.

\begin{figure*}[t]
  \centering
  \includegraphics[width=16cm, height=8cm]{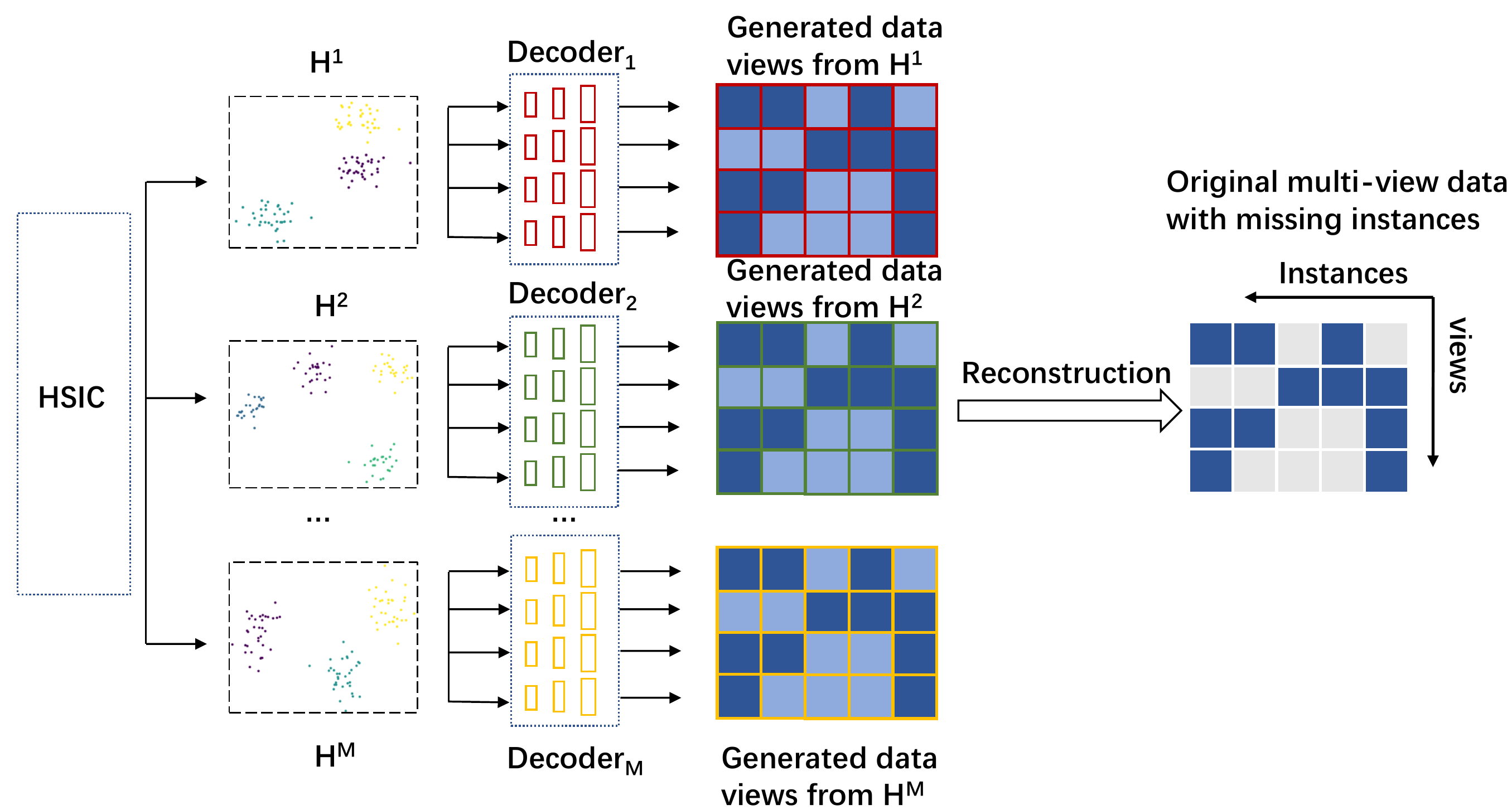}
  \caption{Main schema of DiMVMC to generate alternative clusterings from multi-view data with missing instances (gray blocks on the right). With the assumption that each view is conditionally independent given the shared multi-view representation $P(\mathcal{S}|\mathbf{H}^m)$ ($\mathcal{S}$ stores the observed data views with missing instances), DiMVMC initializes a group of shared subspaces $\{\mathbf{H}^m\}_{m=1}^M$, and then reconstructs the observed missing data views from $\mathbf{H}^m$ via Decoder$_m$.  After the reconstruction, $M$ representational subspaces   $\{\mathbf{H}^m\}_{m=1}^M$ are simultaneously produced, and the features of missing instances are completed. It further uses the HSIC (Hilbert-Schmidt Independence Criterion) to reduce the overlap between these subspaces and thus to generate diverse clusterings.}
  \label{fig:case}
\end{figure*}

To address the drawbacks mentioned above, we propose a deep incomplete multi-view multiple clusterings framework (DiMVMC, as illustrated in Figure \ref{fig:case}). DiMVMC adopts $M$ decoder networks to generate $M$ clusterings from $M$ representational subspaces and to complete the missing features of instances. The input for the $m$-th decoder network is the $m$-th shared subspace $\mathbf{H}^{m}$ which is randomly initialized at first, while the output is the reconstruction of multi-view data. By alternatively optimizing the decoder deep networks and $\{\mathbf{H}^{m}\}_{m=1}^{M}$, DiMVMC can achieve the completeness and $M$ individually shared subspaces $\{\mathbf{H}^{m}\}_{m=1}^{M}$ simultaneously. Moreover, these decoder networks are not isolated, but additionally controlled by a redundancy term based on Hilbert Schmidt Independence Criterion (HSIC) \cite{gretton2005measuring}, which further enforces the diversity among subspaces and thus reduces the redundancy between clusterings. The main contributions of our work are summarized as follows:
\begin{enumerate}[(i)]
\item We study how to generate multiple clusterings on multi-view data with missing samples, which is an important and practical topic, but more challenging and mostly overlooked by previous solutions. To our knowledge, DiMVMC is the first deep approach to generate \textbf{multiple clusterings}.
\item DiMVMC discards the ad-hoc encoder part of autoencoder and works in a unsupervised way, as such DiMVMC has a lower network complexity and can more flexibly deal with data incompleteness in different views. In addition, it uses a redundancy quantification term to reduce the overlap among decoder networks for producing less overlapped representational subspaces, and finally generates diverse clusterings in these subspaces.
\item DiMVMC can find multiple clusterings with higher quality and diversity than the state-of-the-art competitors \cite{jain2008deckmeans,mautz2018NrKMeans,yang2017MNMF,yao2019MVMC,wei2019multi}, and it is robust to missing data in a wide range.
\end{enumerate}

\section{Related Works}
Our work has close connections with two lines of related works, incomplete multi-view clustering and multiple clusterings.
\subsection{Incomplete Multi-view Clustering}
Various multi-view clustering solutions have been introduced, most of which focus on extracting the consistent/complementary information from different views to induce a consolidated clustering \cite{chaudhuri2009multi,kumar2011co-train,wang2015deep}, while others additionally mine the individual information to achieve a more robust clustering \cite{luo2018consistent}. These methods all build on the  assumption that all the data views are complete. While in practice, it is more often that some samples are absent in some views. To handle such more challenging incomplete multi-view clustering (IMC), Li \textit{et al.} \cite{li2014partial} presented the first solution (named PVC) based on NMF (Nonnegative Matrix Factorization) \cite{lee2001NMF}, which learned common representations for complete instances and private latent representations for incomplete instances with the same basis matrices. Next, PVC used the common and private representations to seek a clustering. Zhao \textit{et al.} \cite{zhao2016incomplete} further integrated PVC and manifold learning to learn the global structure of multi-view data. Nevertheless, these NMF-based methods can only deal with two-view data, limiting their application scope. Weighted NMF-based approaches \cite{hu2019doubly,Shao2015multiple} were also proposed to deal with more than two views by filling the missing data and assigning them with lower weights. Wen \textit{et al.} \cite{wen2019unified} added an error matrix to compensate the missing data, and combined the original incomplete data matrix with the error matrix to form a completed data matrix for clustering.  All these solutions in essence build on NMF, which performs shallow projection that cannot well mine the complex relationships between low-level features of multi-view data.

To mine nonlinear structures and complex correlations among multi-view data, Wang \textit{et al.} \cite{wang2018partial} proposed the consistent GAN for the two-view IMC problem, which used one view to generate the missing data of the other view, and then performed clustering on the generated complete data. Xu \textit{et al.} \cite{xu2019adversarial} sought the common latent space of multi-view data and performed missing data inference via combining GAN with autoencoder. Ngiam \textit{et al.} \cite{ngiam2011multimodal}  extracted shared representations by training a two-view deep autoencoder to best reconstruct the two-view inputs. Zhang \textit{et al.} \cite{zhang2019cpm} combined auto-encoder with Bayesian framework to fully exploit partial multi-view data to produce a structured representation for classification. These shallow/deep multi-view clustering solutions still focus on producing a single clustering. Given the multiplicity of multi-view data, it is more desirable    to find different clustering results from the same data and each clustering gives an independent grouping of the data.

\subsection{Multiple Clusterings}
Multiple clusterings focus on how to generate different clusterings with both high quality and diversity from the same data \cite{bailey2013alternative}. It is less well studied than single/multi-view clustering and  ensemble clustering \cite{jain2010data,zhou2012ensemble}, due to its requirement on generating multiple groups of results, and the difficulties on guaranteeing the good quality and diversity at the same time.
Bae \textit{et al.} \cite{bae2006coala} proposed a multiple clusterings solution based on hierarchical clustering (COALA). The main idea of COALA is that instances with higher intra-class similarity still gather in one cluster, while those with lower intra-class similarity are considered to be placed into different clusters for another clustering. Jain \textit{et al.} presented Dec-kmeans \cite{jain2008deckmeans}, which  obtained diverse clusterings simultaneously by finding multiple groups of mutually orthogonal cluster centroids. Unlike COALA and Dec-kmeans that directly control the diversity between clustering results, other solutions   control the diversity between clustering subspaces and then generate different clusterings in these subspaces. Cui \textit{et al.} \cite{cui2007OSC} greedily found orthogonal projection matrices to get different feature representations of the original data and then found clusterings in these orthogonal subspaces. Mautz \textit{et al.} \cite{mautz2018NrKMeans} also tried to explore multiple mutually orthogonal subspaces, along with the optimization of k-means objective function, to find non-redundant clusterings. However, the orthogonal constraint is too strict to generate more than two clusterings.  Wang \textit{et al.} \cite{wang2019MISC} generated multiple independent subspaces with semantic interpretation via independent subspace analysis, and then performed kernel-based clustering in these subspaces to explore diverse clusterings.
Yang and Zhang \cite{yang2017MNMF} explicitly defined a regularization term to quantify and minimize the redundancy between the already generated clusterings and the to-be-generated one, and then plugged this regularization into the matrix factorization based clustering \cite{ding2010convex} to find another   clustering. Wang \textit{et al.} \cite{wang2018mcc} and Yao \textit{et al.}\cite{yao2019mccss} directly minimized the redundancy between all the to-be-generated  clusterings to simultaneously find alternative clusterings. Besides, Caruana \textit{et al.} \cite{caruana2006meta} firstly generated a number of useful high-quality clusterings, and then grouped these clusterings at the meta-level, and thus allowed the user to select a few high-quality and non-redundant clusterings for examination. However, these multiple clusterings methods are still restricted to \textbf{single-view} data.

Given the multiplicity of multi-view data, it is desirable but more difficult to generate multiple clusterings from the same \textbf{multi-view} data. Two approaches have been proposed for attacking this challenging task. MVMC \cite{yao2019MVMC} first extends multi-view self-representation learning \cite{luo2018consistent} to explore the individuality information encoding matrices and the commonality information matrix shared across views, and then combines each individuality similarity matrix and the commonality similarity to generate a distinct clustering by matrix factorization. However, given the cubic time complexity  of the self-representation learning, MVMC can hardly be applicable on datasets with a large number of samples. To alleviate this drawback, DMClusts extends the deep matrix factorization \cite{trigeorgis2016dmf,zhao2017dmf} to collaboratively factorize the multi-view data matrices into multiple representational subspaces layer-by-layer, and seeks a different clustering of high quality per layer. In addition, it introduces a new balanced redundancy quantification term to guarantee the diversity among these clusterings, and thus reduces the overlap between the produced clusterings.

The above-mentioned single/multi-view multiple clusterings solutions  assume all instances are complete across views, and project data into linear and shallow subspaces. Therefore, they cannot capture the complex correlations between views and nonlinear clusters in subspaces when data are incomplete. To address these issues, we introduce  DiMVMC to mine multiple clusterings from multi-view data with missing instances. DiMVMC can capture the complex correlations among views and complete data by multiple decoder networks, and thus generate multiple nonlinear clusterings with quality and diversity.

\section{The Proposed Method}
It was empirically demonstrated that different data views are complementary to each other, and they carry distinct information for generating diverse clusterings with quality \cite{yao2019MVMC,wei2019multi}. However, multi-view multiple clusterings is still challenging due to the difficulty in modeling the unknown and complex correlation among different views. Moreover, data with missing views and the required diversity between clusterings further upgrade the difficulty to address the incomplete multi-view multiple clusterings problem. Autoencoder is typically used to reconstruct the data with missing/noisy features \cite{hinton1994AE,xu2019adversarial,zhang2019cpm}.
The encoder takes input the incomplete data and learns a compact representation, from which the decoder recovers the missing values. To avoid the ad-hoc design of encoder for the incomplete cases in different views,
we skip the encoder and take the shared subspace representation $\mathbf{H}^m$ as the input for the $m$-th decoder
network, from which 
the observed data   are reconstructed and  the missing data are completed, as shown in Fig. 1. In addition, we quantify and minimize the redundancy among these subspaces for generating diverse clusterings therein. The following subsections elaborate on the above procedure.



\subsection{Generating Multiple Representation Subspaces}
Suppose a multi-view dataset with $V$ views has $N$ instances. We use $\mathbf{x}_n^{v} \in \mathbb{R}^{d_v}$ ($v=1,\cdots,V$) to denote the feature vector for the $v$-th view of the $n$-th instance, where $d_v$ is the feature dimension of the $v$-th view. An indicator matrix $\mathbf{\Lambda} \in \{0,1\}^{V\times N}$ for all instances is defined as:
\begin{equation}
\small
\mathbf{\Lambda}_{vn}=
\begin{cases}
1& \text{if the $n$-th instance has the $v$-th view}\\
0& \text{otherwise}
\end{cases}
\label{Eq1}
\end{equation}
where each column of $\mathbf{\Lambda}$ is the status (present/absent) of instances for corresponding views. The relation $1 \leq {\sum}_{v=1}^V\mathbf{\Lambda}_{vn}\leq V$ holds, such that each instance is present in at least one view.

The aim of incomplete multi-view multiple clusterings is to integrate all the incomplete views to generate multiple clusterings. Inspired by cross partial multi-view networks for classification and by multi-view subspace learning \cite{zhang2019cpm,white2012convex}, we project instances with arbitrary view-missing patterns into the shared representational subspaces in a flexible way, where the  subspaces include the information for observed views. That is, each view can be reconstructed by the obtained shared representation. Based on the reconstruction point of view \cite{lee1996image}, we use the joint distribution to concrete the above idea as follows:
\begin{equation}
\small
 \begin{split}
P(\mathcal{S}_i|\mathbf{h}_i)=\prod_{v=1}^VP(\mathbf{x}_i^v|\mathbf{h}_i)
\end{split}
\label{Eq2}
\end{equation}
where  $\mathbf{h}_i\in \mathbb{R}^{d}$ is the multi-view shared representation of the $i$-th instance, and $\mathcal{S}_i={\{\mathbf{x}_i^v\}}_{v=1}^V$, By maximizing $P(\mathcal{S}_i|\mathbf{h}_i)$, the common subspaces ${\{\mathbf{h}_i\}}_{i=1}^{N}$ can be obtained. {However, just alike typical subspace learning methods \cite{Vidal2011Subspace,sparsesc},
(\ref{Eq2}) also optimizes one subspace and a single clustering result therein. Because of the multiplexes, multi-view data has a mix of diverse distributions. Therefore, multiple different subspaces and clusterings can co-exist.}
To gain multiple ($M$) clusterings, we extend  (\ref{Eq2}) as:
\begin{equation}
\small
 \begin{split}
P(\mathcal{S}_i|\mathbf{h}_{i}^m)=\prod_{v=1}^VP(\mathbf{x}_i^v|\mathbf{h}_{i}^m)
\end{split}
\label{Eq3}
\end{equation}
where $\mathbf{h}_{i}^m$ is the shared representation of the $i$-th instance in the $m$-th shared subspace.
Based on different views in $\mathcal{S}_i$, we model the likelihood with respect to $\mathbf{h}_{i}^m$ given the observation $\mathbf{x}_i^v$ as:
\begin{equation}
\small
 \begin{split}
P(\mathbf{x}_i^v|\mathbf{h}_{i}^m)\propto{e^{-\Delta(\mathbf{x}_i^v, f_{m}^v(\mathbf{h}_{i}^m, \Theta_m^v))}}
\end{split}
\label{Eq4}
\end{equation}
where $\Delta$ represents the reconstruction loss. Here, we adopt the $l_2$-norm for this reconstruction part. $f_{m}^v$ is the mapping function from the common subspace $\mathbf{H}^m$ to the $v$-th view and $\Theta_m^v$ are decoder network parameters of $f_{m}^v$.

Without loss of generality, suppose the data are independent and identically distributed, we can induce the log-likelihood function as follows:
\begin{equation}
\small
 \begin{split}
\mathcal{L}(\{\mathbf{H}^m\}_{m=1}^M, \{\Theta_m^v\}_{m=1,v=1}^{M,V})=\sum_{m=1}^{M}  \sum_{i=1}^{N}ln P(\mathcal{S}_i|\mathbf{h}_{i}^m)
\end{split}
\label{Eq5}
\end{equation}
Since maximizing the likelihood is equivalent to minimizing the loss $\Delta$, by considering the missing case, we can obtain the following objective function for the decoder network:
\begin{equation}
\small
 \begin{split}
\underset{\{\mathbf{H}^m\}_{m=1}^M, \{\Theta_m^v\}_{m=1,v=1}^{M,V}}{min}\sum_{m=1}^{M} \sum_{v=1}^{V} \sum_{i=1}^{N} \Lambda_{vi}\Delta(\mathbf{x}_i^v, f_{m}^v(\mathbf{h}_i^m, \Theta_m^v))
\end{split}
\label{Eq6}
\end{equation}
Optimizing the above equation can generate $M$ shared representations $\{\mathbf{H}_m\}_{m=1}^M$, each of which will be used to generate a  clustering result. Unlike traditional autoencoder based solutions \cite{xu2019adversarial,zhang2019cpm}, DiMVMC skips the encoder networks, but takes the shared subspace representation $\mathbf{H}^m$ as the input for the $m$-th decoder to complete multi-view data, as done by $f_{m}^v$ in (\ref{Eq6}). As such, DiMVMC does not need to specifically consider diverse missing cases of multi-view data, while still makes full use of observed data.

\subsection{Reducing Redundancy between Subspaces}
By minimizing (\ref{Eq6}), we can generate multiple common subspaces from incomplete multi-view data. For multiple clusterings, besides the quality of different clusterings, the diversity between clusterings is also important \cite{bailey2013alternative}. The diversity is usually approximately obtained by minimizing the redundancy between these subspaces. Orthogonality is the most common approach that forces two subspaces being orthogonal with each other. Orthogonality based methods may still generate multiple clusterings with high redundancy, since these orthogonal subspaces can still produce clusters with the same structure \cite{wei2019multi}. Furthermore, orthogonality does not specify which properties of the reference clustering should or should not be retained. Kullback Leibler (KL) divergence was also adopted to find diverse clusterings \cite{qi2009principled}, but KL divergence is not symmetric and not applicable for high-dimensional data, due to its high time and space complexity.

Hilbert-Schmidt Independence Criterion (HSIC) \cite{gretton2005measuring} measures the squared norm of the cross-covariance operator over $\mathbf{H}^m$ and $\mathbf{H}^{m'}$ in the Hilbert kernel space to estimate the dependency. It is empirically given by:
\begin{equation}
\small
\begin{split}
HSIC(\mathbf{H}^m, \mathbf{H}^{m'})=\frac{1}{(N-1)^{2}}tr(\mathbf{K}^m \mathbf{A} \mathbf{K}^{m'}\mathbf{A})
\end{split}
\label{Eq8}
\end{equation}
where $\mathbf{K}^m$ and $\mathbf{K}^{m'} \in \mathbb{R}^{N\times N}$ are Gram matrices, defined as an inner product between vectors in a specific kernel space. $\mathbf{A}_{ij}=\delta_{ij}-1/N$, $\delta_{ij}=1$ if $i=j$, $\delta_{ij}=0$ otherwise.  In this paper, we adopt the inner product kernel to specify $\mathbf{K}^m=(\mathbf{H}^m)^T\mathbf{H}^m$. A lower HSIC value means two subspaces are less correlated. This empirical estimation is simpler than any other kernel dependence test, and requires no user-defined regularisation. In addition, it has a solid theoretical foundation, a fast learning rate with guaranteed exponential convergence, and the capability in measuring both linear and nonlinear dependence between variables. For these merits, we adopt HSIC to quantify the overlap between generated subspaces $\{\mathbf{H}^m\}_{m=1}^M$.

\subsection{Unified Model}
By integrating {(\ref{Eq6}) with (\ref{Eq8})}, we define the loss function of DiMVMC as:
\begin{equation}
\small
\begin{split}
\mathcal{J}_1(\Theta_m^v, \mathbf{H}^m)&=\Phi\sum_{m=1}^{M}\sum_{v=1}^{V}\sum_{i=1}^{N}\Lambda_{vi} \Delta(\mathbf{x}_i^v, f_{m}^v(\mathbf{h}_{i}^m, \Theta_m^v))+\\
&\lambda\sum_{m=1,m\neq m'}^{M}HSIC(\mathbf{H}^m, \mathbf{H}^{m'})
\end{split}
\label{Eq9}
\end{equation}
where $\Phi=\frac{1}{N^2d_{ave}^2}$ ($d_{ave}$ is the average of $d_v$, $v=1, 2, \cdots, V$) is the normalization factor, $\lambda$ is the hyper-parameter to balance the sought of $M$ subspaces and diversity between them. DiMVMC can generate multiple common subspaces $\{\mathbf{H}^m\}_{m=1}^M$ and complete missing data simultaneously via minimizing (\ref{Eq9}). Since the optimal solution cannot be analytically given, we employ an optimization strategy that alternatively updates $\Theta_m^v$ or $\mathbf{H}^m$ in an iterative way, while fixing the others. More specifically, $\Theta_m^v$ and $\mathbf{H}^m$ are randomly initialized at first.  The detailed optimization process is given in Algorithm \ref{alg:1}. Once the optimization is done, $k$-means is implemented on each obtained subspace $\mathbf{H}^m$, and thus $M$ clusterings with quality and diversity can be accordingly generated.

\begin{algorithm}[t]
\caption{DiMVMC: Deep incomplete Multi-view Multiple Clusterings}
\label{alg:1}
{\bf Input:} Incomplete multi-view dataset $\mathcal{X}$, scalar parameters $\lambda$, number of subspaces $M$, learning rate $\eta$.\\
{\bf Output:} Networks parameters $\{\Theta_m^v\}_{m=1,v=1}^{M,V}$,  $M$ subspaces $\{\mathbf{H}^m\}_{m=1}^M$ and $M$ alternative clusterings $\{\mathcal{C}^m\}_{m=1}^M$ in these subspaces.
\begin{algorithmic}[1]
\STATE Random initialization for $\{\mathbf{H}^m\}_{m=1}^M$ and $\{\Theta_m^v\}_{m=1,v=1}^{M,V}$
\WHILE{not converged}
    \FOR{$v=1\colon V$}
        \FOR{$m=1\colon M$}
            \STATE $\Theta_m^v\leftarrow\Theta_m^v-\eta\partial\mathcal{J}/\partial\Theta_m^v$;
        \ENDFOR
    \ENDFOR
    \FOR{$m=1\colon M$}
       \STATE $\mathbf{H}^m\leftarrow\mathbf{H}^m-\eta\partial\mathcal{J}/\partial\mathbf{H}^m$;
    \ENDFOR
\ENDWHILE\\
\STATE Grouping all instances via $k$-means in the representational subspaces $\{\mathbf{H}^m\}_{m=1}^M$.
\end{algorithmic}
\end{algorithm}

In the subspace clustering, it is desired that the subspace representation is sparse but captures the group-level semantic information. There are different ways to bring in sparsity. For example, we can add a drop-out layer for the deep learning approach. Here, to make an intuitive implementation, we add a sparsity-induced regularization into the above loss function and define a so-called Sparse DiMVMC:
\begin{equation}
\small
 \begin{split}
\mathcal{J}_2(\Theta_m^v, \mathbf{H}^m)&=\Phi\sum_{m=1}^{M}\sum_{v=1}^{V}\sum_{i=1}^{N}\Lambda_{vi} \Delta(\mathbf{x}_i^v, f_{m}^v(\mathbf{h}_{i}^m, \Theta_m^v))+\\
&\lambda\sum_{m=1,m\neq m'}^{M}HSIC(\mathbf{H}^m, \mathbf{H}^{m'})+\alpha\sum_{m=1}^{M}||\mathbf{H}^m||_1
\end{split}
\label{Eq10}
\end{equation}
where $||\cdot||_1$ is the $l_1$ norm of the matrix. When $\alpha = 0$, (\ref{Eq10}) goes back to the plain DiMVMC.

\section{Experimental Results and Analysis}

\subsection{Experimental Setup}
In this section, we evaluate the performance of our proposed DiMVMC on four benchmark multi-view datasets, as described in Table \ref{table1}. \textbf{Caltech20} \cite{li2015large} is a subset of Caltech-101 of 6 categories, which contains 2,386 instances and 20 clusters. We utilize 254D CENHIST vector, 512D GIST vector and 928D LBP vector as three views. \textbf{Handwritten} \cite{li2015large} is comprised of 2,000 data points from 0 to 9 digit classes, with 200 data points for each class. There are six public features available. We utilize 240D pixel averages feature, 216D profile correlations feature and 76D LBP feature as three views. \textbf{Reuters} \cite{NIPS2009_3690} is a textual data set consisting of 111,740 documents in five different languages (English, French, German, Spanish and Italian) of 6 classes. We randomly sample 12000 documents from this collection in a balanced manner. We further do dimensionality reduction on the 12000 samples following the methodology of \cite{cotrmv2011} and represent each view by 256 numeric features.  \textbf{Mirflickr} includes 25,000 samples collected from Flicker with an image view and a textual view. Here, we remove textual tags that appear less than 20 times, and then remove samples without textual tags or semantic labels, finally we get 16,738 samples for experiments \cite{yao2019MVMC}.

\begin{table}[h!tbp]
\scriptsize
		\caption{Statistics of used multi-view datasets. $N$, $V$ and $K$ are the numbers of instances, views and clusters. $\{d_v\}_{v=1}^V$ are the feature dimensions of $V$ views.}
		\centering
\begin{tabular}{c|r|r |c}
	\hline
	Datasets &$N$, $V$ &$K$ &$d_v$\\
	\hline
    Caltech20 &2386, 3 &20 &[254, 512, 928]\\
    Handwritten &2000, 3 &10 &[240, 216, 76]\\
    Reuters &12000, 5 &6 &[256, 256, 256, 256, 256]\\
    Mirflickr &16738, 2 &24 &[150, 500]\\
	\hline
\end{tabular}
\label{table1}
\end{table}




We compare DiMVMC against with six representative and recent multiple clusterings algorithms, including Dec-kmeans \cite{jain2008deckmeans}, Nr-kmeans \cite{mautz2018NrKMeans}, OSC \cite{cui2007OSC}, MNMF \cite{yang2017MNMF}, MVMC \cite{yao2019MVMC} and DMClusts \cite{wei2019multi}. Dec-kmeans is a representative multiple clusterings solution based on orthogonalizing the clustering centroids. Nr-kmeans \cite{mautz2018NrKMeans}, OSC \cite{cui2007OSC} and MNMF \cite{yang2017MNMF} attempt different techniques to seek subspaces and multiple clusterings therein. For this reason, they have close connections with our approach and are used for experimental comparison. MVMC \cite{yao2019MVMC} and DMClusts \cite{wei2019multi} are the only two multi-view multiple clusterings algorithms at present.

None of these compared multiple clusterings algorithms can directly handle missing data, we fill the missing features (instances) with average values for each view at first, and then apply these solutions. For those single-view algorithms, following the solution in \cite{yao2019MVMC,wei2019multi}, we concatenate the feature vectors of multi-view data and then apply them on the concatenated vectors to generate different clusterings. Note, we do not take the deep/incomplete multi-view clustering solutions for experiments, since they can only output a single clustering. In the following experiments, the input parameters of the comparing methods are fixed (or optimized) as the authors suggested in their papers or generously shared codes.

DiMVMC selected the input parameters from the following ranges: $\lambda \in \{10^{-3}, 10^{-2}, \cdots, 10^{3}\}$, $d \in \{64, 128\}$, and $M=2$. The alternate clusterings are generated by applying $k$-means algorithm on each shared subspace $\mathbf{H}^m$. The number of clusters for alternate clusterings was fixed to $K$ of each dataset, as listed in Table \ref{table1}. In this paper, DiMVMC simply adopts two layers' network structure for each mapping $f_m^v$ to reconstruct the $v$-th view from the $m$-th shared subspace $\mathbf{H}^m$. {The demo code of DiMVMC is available at \href{http://mlda.swu.edu.cn/codes.php?name=DiMVMC}{http://mlda.swu.edu.cn/codes.php?name=DiMVMC}}.


Following the evaluation protocol used by the baseline methods \cite{yang2017MNMF,wang2018mcc,yao2019MVMC}, we measure the \emph{quality} of multiple clusterings via the average SC (Silhouette Coefficient) or DI (Dunn Index), and the \emph{diversity} via the NMI (Normalized Mutual Information) or JC (Jaccard Coefficient).  SC and DI quantify the compactness and separation of clusters within a clustering, while NMI and JC  quantify the similarity of clusters of two clusterings $\mathcal{C}^1$ and $\mathcal{C}^2$. We want to remark that unlike traditional clustering problem, a lower value of NMI and JC means the two alternative cluterings are less overlapped, so a smaller value of them is more preferred.

\subsection{Discovering Multiple Clusterings}
\begin{table*}[t]
	\caption{Quality and Diversity of  various compared methods on generating multiple clusterings from `complete' multi-view datasets. $\uparrow$($\downarrow$) indicates the preferred direction for the corresponding evaluation metric. $\bullet / \circ$ indicates whether our DiMVMC is statistically (according to pairwise $t$-test at 95\% significance level) superior/inferior to the other method.}
    \centering
	\resizebox{1.9\columnwidth}{!}{
	\begin{tabular}{c|c| r r r r r r |r}
		\hline
		& &Dec-kmeans &Nr-kmeans &OSC &MNMF &MVMC &DMClusts  &DiMVMC\\
		\hline
        \multirow{4}[2]{*}{Caltech20}
        &SC$\uparrow$
        & -0.107$\pm$0.002$\bullet$ & 0.053$\pm$0.003$\circ$ & 0.190$\pm$0.001$\circ$ &-0.109$\pm$0.003$\bullet$ & -0.097$\pm$0.001$\bullet$ & -0.033$\pm$0.002$\bullet$ & 0.006$\pm$0.000\\
        &DI$\uparrow$
        & 0.032$\pm$0.000$\bullet$ & 0.042$\pm$0.000$\bullet$ & 0.045$\pm$0.002$\bullet$ & 0.026$\pm$0.000$\bullet$ & 0.009$\pm$0.000$\bullet$ & 0.115$\pm$0.003$\bullet$ & 0.265$\pm$0.006  \\
        &NMI$\downarrow$
        & 0.053$\pm$0.002$\bullet$ & 0.465$\pm$0.011$\bullet$ & 0.667$\pm$0.007$\bullet$ & 0.070$\pm$0.001$\bullet$ & 0.025$\pm$0.003$\bullet$ & 0.065$\pm$0.002$\bullet$ & 0.024$\pm$0.001  \\
        &JC$\downarrow$
        & 0.046$\pm$0.000$\bullet$ & 0.176$\pm$0.008$\bullet$ & 0.297$\pm$0.004$\bullet$ & 0.045$\pm$0.000$\bullet$ & 0.026$\pm$0.001$\bullet$ & 0.050$\pm$0.001$\bullet$ & 0.025$\pm$0.001  \\
        \hline
        \multirow{4}[2]{*}{Handwritten}
        &SC$\uparrow$
        & 0.043$\pm$0.001$\circ$ & 0.126$\pm$0.004$\circ$ &0.371$\pm$0.003$\circ$ &0.020$\pm$0.001$\circ$ &-0.024$\pm$0.000$\bullet$ &0.020$\pm$0.001$\circ$ &0.007$\pm$0.000 \\
        &DI$\uparrow$
        & 0.056$\pm$0.000$\bullet$ & 0.068$\pm$0.001$\bullet$ &0.074$\pm$0.002$\bullet$ &0.031$\pm$0.002$\bullet$ &0.004$\pm$0.000$\bullet$ &0.173$\pm$0.006$\bullet$ &0.604$\pm$0.009 \\
        &NMI$\downarrow$
        & 0.057$\pm$0.001$\bullet$ & 0.395$\pm$0.010$\bullet$ &0.756$\pm$0.003$\bullet$ &0.093$\pm$0.001$\bullet$ &0.006$\pm$0.001 &0.061$\pm$0.005$\bullet$ &0.006$\pm$0.000 \\
        &JC$\downarrow$
        & 0.065$\pm$0.001$\bullet$ & 0.207$\pm$0.011$\bullet$ &0.637$\pm$0.020$\bullet$ &0.078$\pm$0.002$\circ$ &0.091$\pm$0.001$\bullet$ &0.095$\pm$0.003$\bullet$ &0.052$\pm$0.001 \\
        \hline
        \multirow{4}[2]{*}{Reuters}
        &SC$\uparrow$
        & -0.033$\pm$0.000$\bullet$ & -0.012$\pm$0.000$\bullet$ &0.013$\pm$0.001$\circ$ &-0.002$\pm$0.000$\bullet$ &-0.004$\pm$0.000$\bullet$ &0.015$\pm$0.001$\circ$ &0.011$\pm$0.000  \\
        &DI$\uparrow$
        & 0.047$\pm$0.002$\bullet$ & 0.055$\pm$0.003$\bullet$ &0.068$\pm$0.002$\bullet$ &0.019$\pm$0.001$\bullet$ &0.013$\pm$0.000$\bullet$ &0.030$\pm$0.001$\bullet$ &0.434$\pm$0.000\\
        &NMI$\downarrow$
        & 0.231$\pm$0.003$\bullet$ & 0.301$\pm$0.005$\bullet$ &0.236$\pm$0.011$\bullet$ &0.001$\pm$0.000 &0.001$\pm$0.000 &0.007$\pm$0.001$\bullet$ &0.001 $\pm$0.000\\
        &JC$\downarrow$
        & 0.290$\pm$0.005$\bullet$ & 0.284$\pm$0.012$\bullet$ & 0.339$\pm$0.009$\bullet$ &0.094$\pm$0.000$\bullet$ &0.093$\pm$0.000$\bullet$ &0.114$\pm$0.003$\bullet$ &0.091 $\pm$0.000\\
        \hline
         \multirow{4}[2]{*}{Mirflickr}
        &SC$\uparrow$
        & -0.004$\pm$0.000$\bullet$ & 0.001$\pm$0.000$\bullet$ &0.017$\pm$0.000$\circ$ &-0.058$\pm$0.000$\bullet$ &-0.038$\pm$0.000$\bullet$  &0.336$\pm$0.008$\circ$ &0.006$\pm$0.000 \\
        &DI$\uparrow$
        &0.061$\pm$0.002$\bullet$ & 0.035$\pm$0.001$\bullet$
         &0.059$\pm$0.002$\bullet$ &0.053$\pm$0.001$\bullet$ &0.173$\pm$0.005$\bullet$ &0.076$\pm$0.001$\bullet$ &0.536$\pm$0.013\\
        &NMI$\downarrow$
        &0.427$\pm$0.012$\bullet$ & 0.584$\pm$0.005$\bullet$ &0.575$\pm$0.011$\bullet$ &0.014$\pm$0.000$\bullet$ &0.005$\pm$0.000 &0.043$\pm$0.001$\bullet$ &0.005$\pm$0.000 \\
        &JC$\downarrow$
        &0.878$\pm$0.022$\bullet$ & 0.363$\pm$0.007$\bullet$ &0.368$\pm$0.011$\bullet$ &0.023$\pm$0.000$\bullet$ &0.022$\pm$0.000$\bullet$ &0.033$\pm$0.001$\bullet$ &0.021$\pm$0.000 \\
        \hline
\end{tabular}}
\label{table2}
\end{table*}

For the first experiment, we assume the four multi-view datasets are complete without any missing data. We report the average results of ten independent runs and standard deviations of each method on generating two alternative clusterings in Table \ref{table2}.

From Table \ref{table2}, we has the following important observations:\\
(i) \textbf{Multi-view vs. Single-view}: DiMVMC, DMClusts and MVMC can be directly applied on multi-view data, and their generated two clusterings often have a lower redundancy than those generated by other compared methods. That is because these methods lack a redundancy control term or their redundancy strategies are difficult to be optimized. Our following experiment will further analyze the importance of redundancy control. DiMVMC frequently obtains a better quality than compared methods that can only work on the concatenated single view, which suggests that concatenating the feature vectors overrides the intrinsic nature of multi-view data, which helps to generate multiple clusterings with quality. These comparisons prove the effectiveness of our proposed approach in fusing multiple data views to generate multiple clusterings with diversity and quality.\\
(ii) \textbf{Shallow methods vs. DiMVMC}:
To our knowledge, DiMVMC is the first deep approach to generate multiple clusterings, and it often performs better on quality metrics (SC and DI), owing to the high-level expression ability of decoder networks. Even though, DiMVMC sporadically has a lower value on SC than some of compared methods. That is due to the widely-recognized dilemma of obtaining alternative clusterings with both high diversity and quality. DiMVMC has a larger diversity. That is explainable, since it can explore diverse nonlinear representation subspaces by decoder networks, while these shallow methods can only obtain low-level feature subspaces. Therefore, DiMVMC has a better tradeoff between quality and diversity than these compared methods. Although DiMVMC, DMClusts and MVMC  can generate diverse clusterings from the same multi-view data, DiMVMC manifests a better performance than the latter two. That is because DiMVMC can mine the complex correlations between views and features via decoder networks, whereas these compared methods cannot.


In summary, even with complete data across views, DiMVMC outperforms compared methods across different multi-view datasets in terms of quality and diversity.
\subsection{Impact of Missing Data}
\begin{figure*}[t]	
	\centering
	\begin{subfigure}{0.96\columnwidth}
		\centering
		\includegraphics[width=8cm, height=5.5cm]{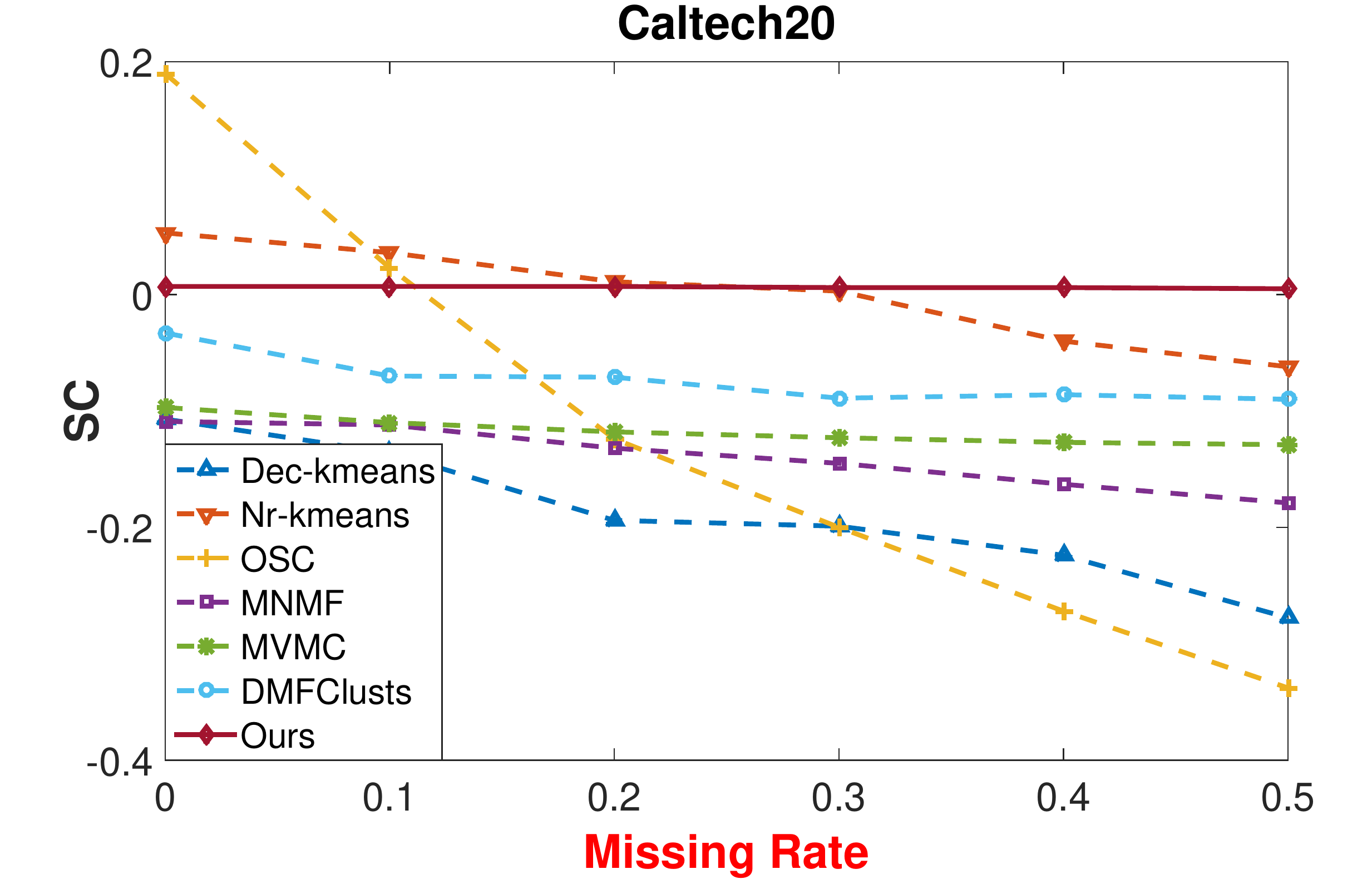}\\
        \includegraphics[width=8cm, height=5.5cm]{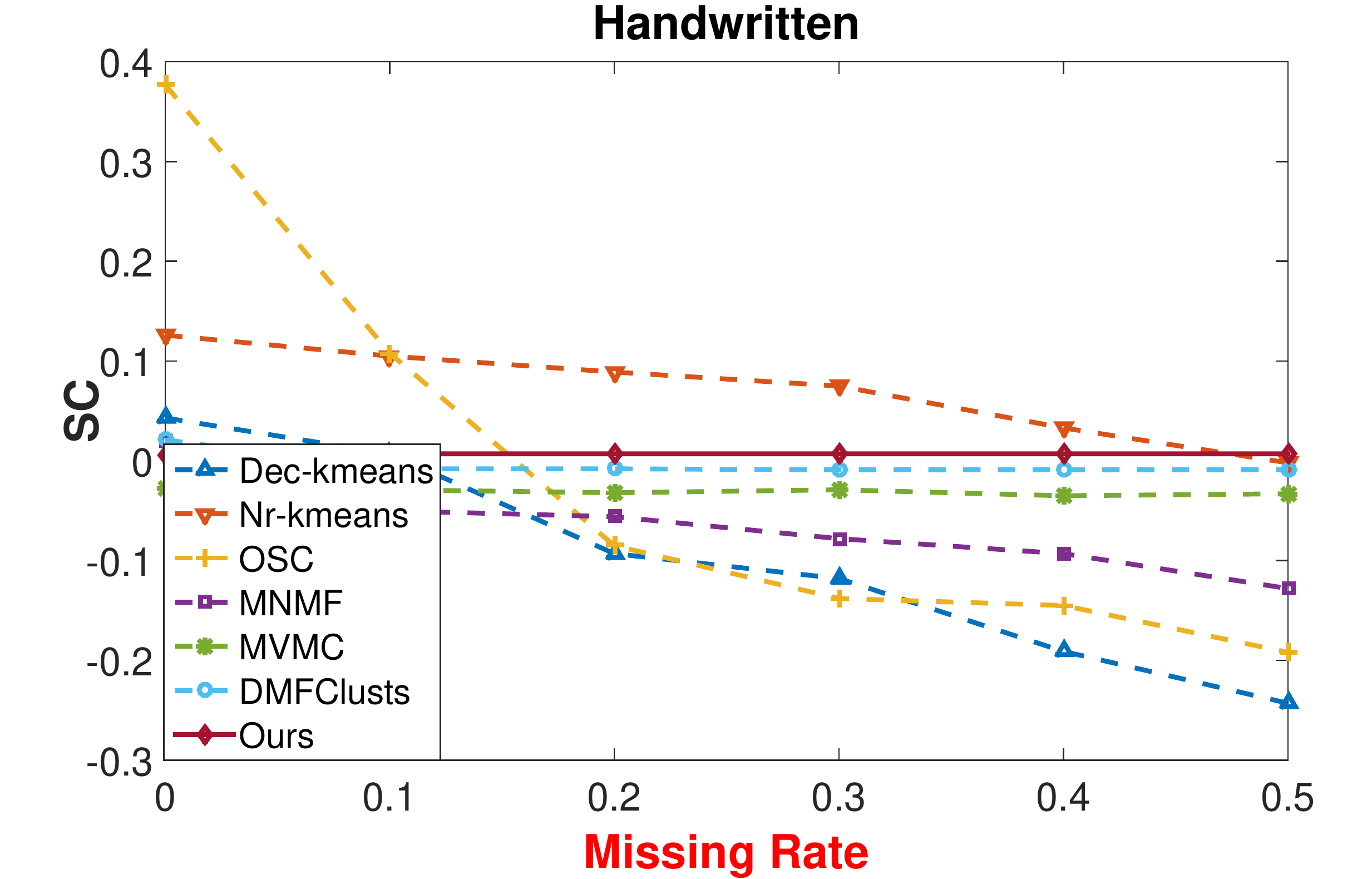}\\
        \includegraphics[width=8cm, height=5.5cm]{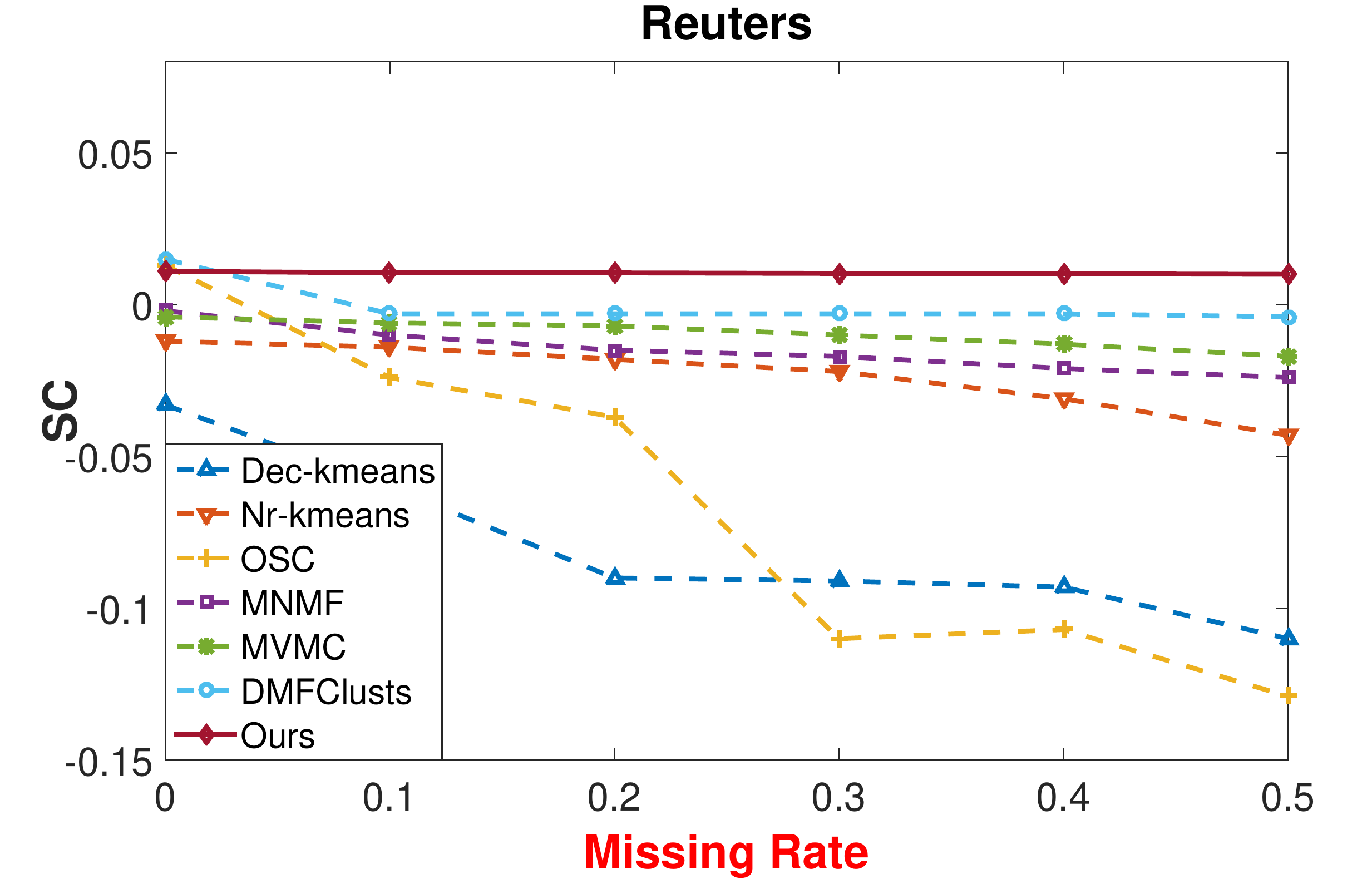}
        \caption{SC}
	\end{subfigure}
	\quad
	\begin{subfigure}{0.96\columnwidth}
		\centering
		\includegraphics[width=8cm, height=5.5cm]{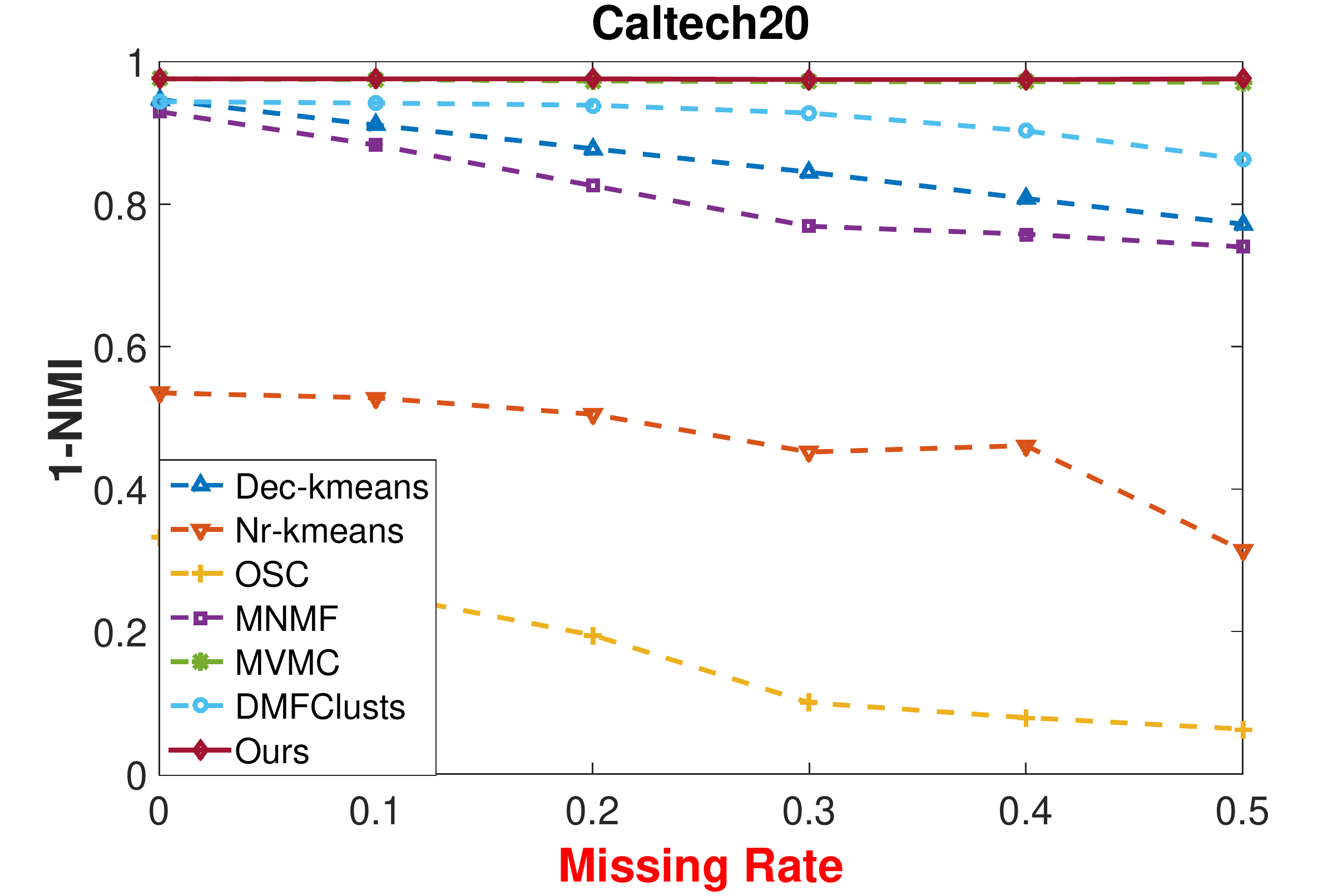}\\
        \includegraphics[width=8cm, height=5.5cm]{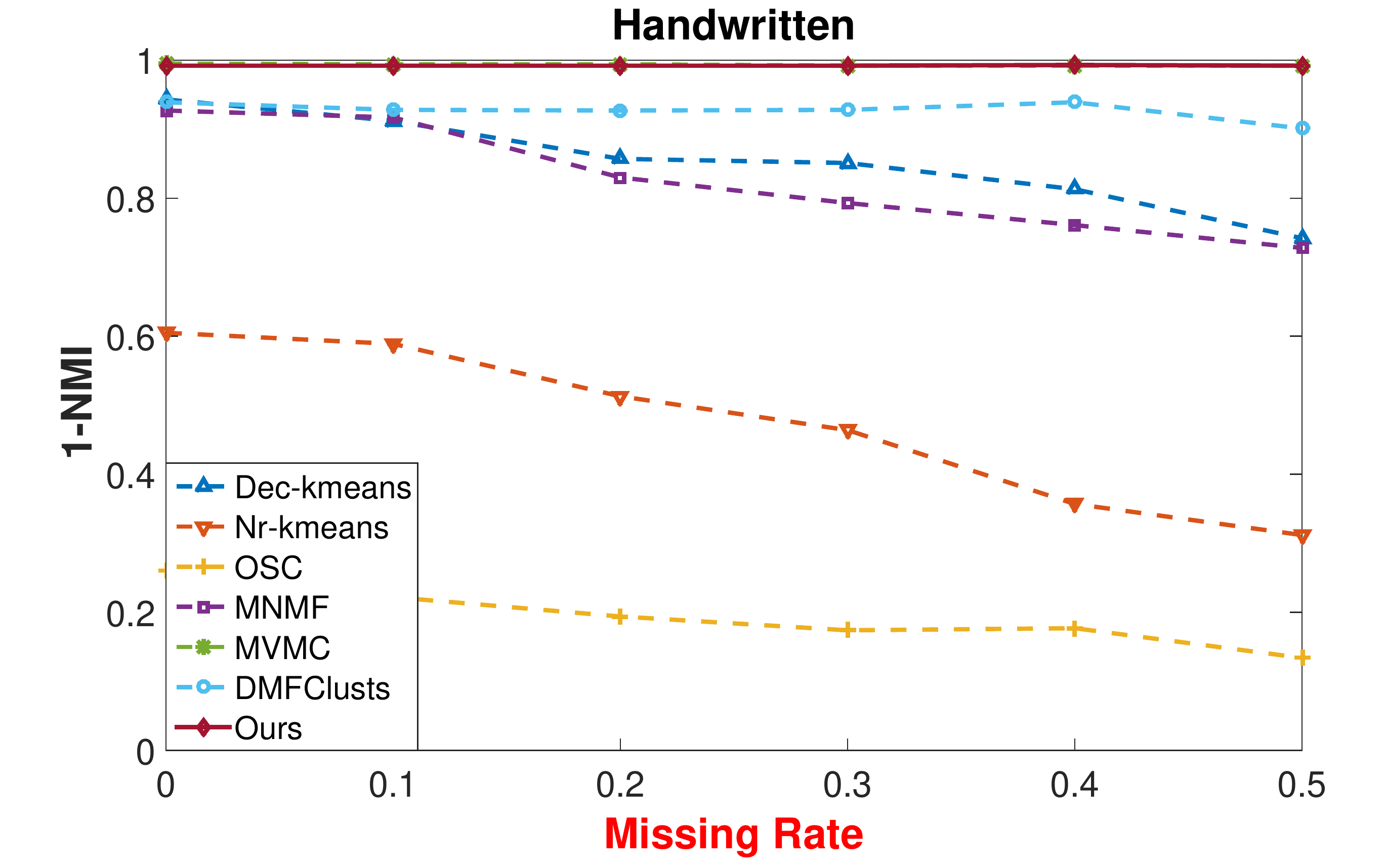}\\
        \includegraphics[width=8cm, height=5.5cm]{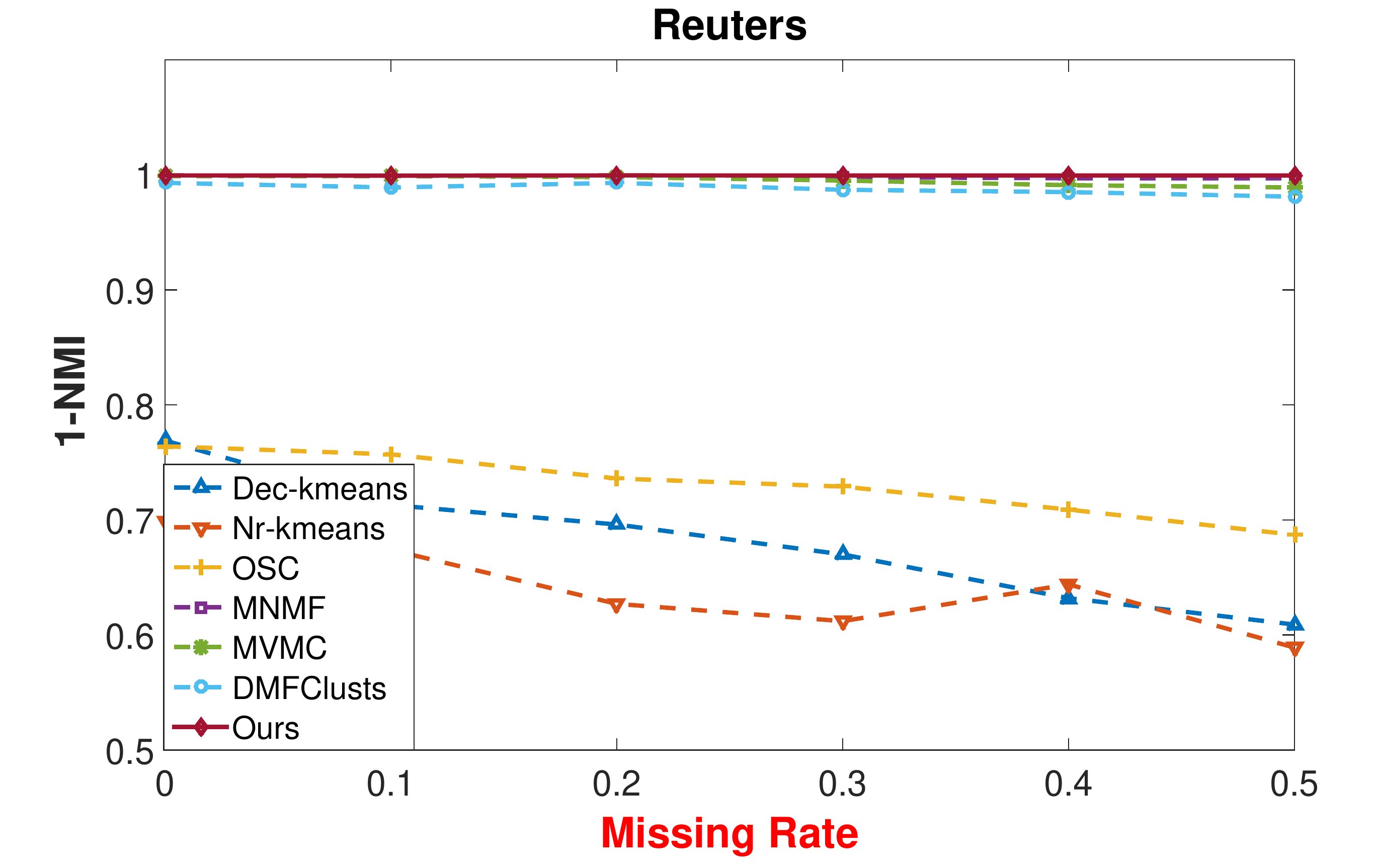}
        \caption{1-NMI}
	\end{subfigure}
	\caption{The variation of quality (SC) and diversity (1-NMI) as the missing rate $\tau$ change.}
    \label{fig2}
\end{figure*}
To study  the performance of DiMVMC with missing data views, we define the missing rate as $\tau=1-\frac{\sum_{v,n}\mathbf{\Lambda}_{vn}}{V\times N}$. The instances are randomly selected as missing ones, and the missing views are randomly erased by guaranteeing at least one of them is observed. In this paper, the missing rate is varied from 0 to 50\% with an interval as 10\%.

Figure \ref{fig2} shows the impact of missing rate of data on the clustering performance of DiMVMC and of compared methods. With the increase of missing rate, the performance of multiple clusterings methods does degrade. Nr-kmeans \cite{mautz2018NrKMeans} and OSC \cite{cui2007OSC} are always in a high position in terms of SC at the beginning. This indicates that they can obtain multiple clusterings of high quality under a small rate of missing instances. However, their SC values drop faster than others with the further increase of missing rate, since they do not take into account the missing instances/features. Furthermore, their diversity (1-NMI) is also at a low-level, suggesting their orthogonal subspaces still have a relatively high redundancy.

The performance curves of multi-view  methods (MVMC, DMClusts and DiMVMC) drop more slowly than the single-view methods as the increase of missing rate. That is because the correlation between views helps to reduce the impact of missing data, and concatenating features cannot well capture this complementary information. In addition, although the SC curve of DiMVMC is not always in a relatively high level, it always holds better diversity than compared methods. In addition, it holds more stable  quality (SC) and diversity (1-NMI) curves than compared methods. This observation again echoes the dilemma of balancing the quality and diversity of multiple clusterings.

Finally,  {DiMVMC can generate clusterings of diversity controlled by the HSIC term.} It is the first multiple clusterings algorithm that considers the missing data views, it can reconstruct the incomplete multi-view dataset to complete the missing data views. As such, DiMVMC is more robust to missing data. By contrast, the simple data complement strategies used by compared methods are not so robust. As a result, the performance of the compared methods is not as stable as DiMVMC is.

Figure \ref{fig2} proves that DiMVMC has a better tradeoff between the quality and diversity of multiple clusterings, and is more competent in dealing with incomplete data than compared methods. That can be attributed to the adopted decoder networks and diversity control term, which can more well capture the correlations among different views and handle diverse missing patterns, and enforce the diversity among subspaces.

\begin{figure*}[t]	
	\centering
	\begin{subfigure}{0.94\columnwidth}
		\centering
		\includegraphics[width=1\columnwidth]{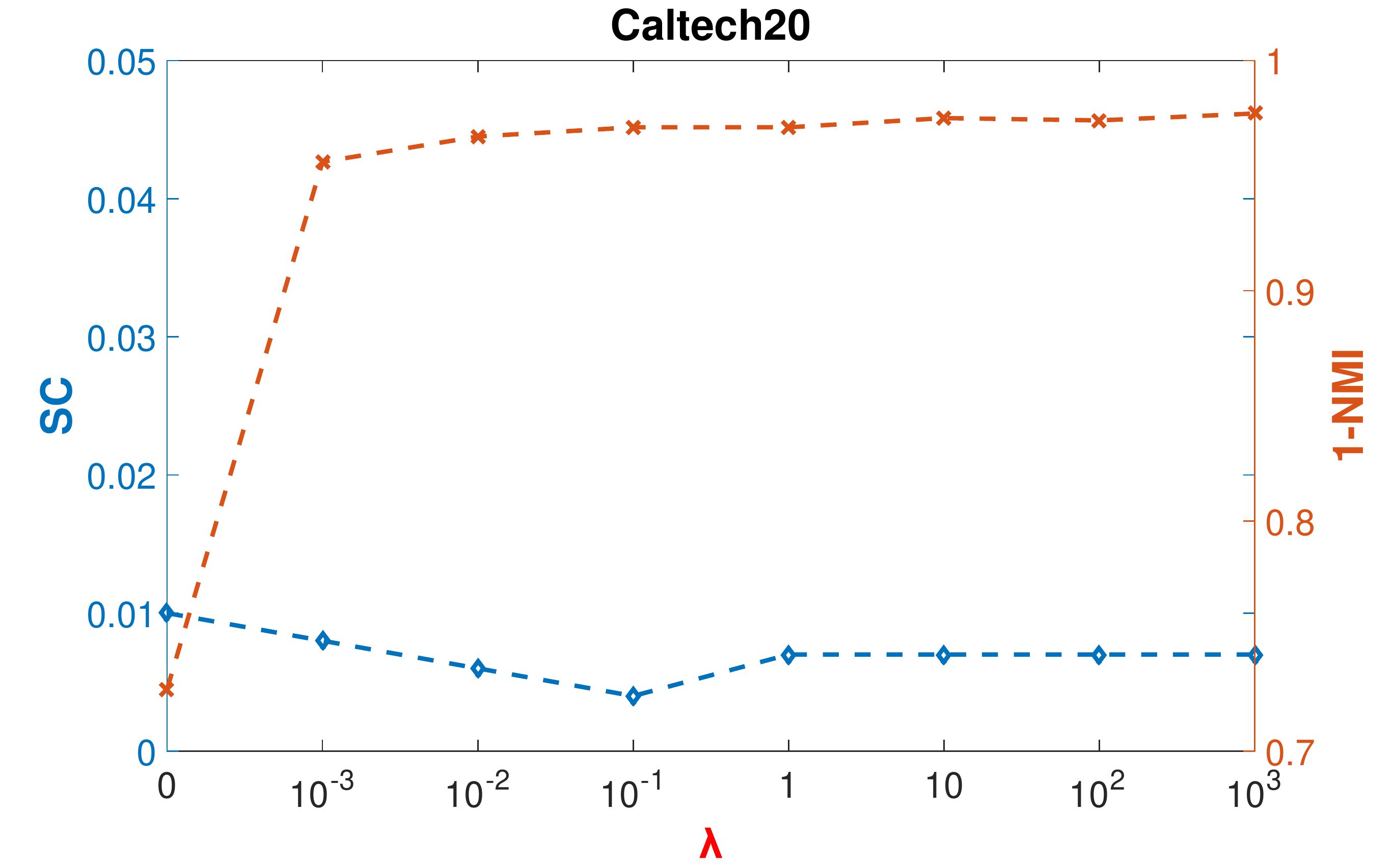}
        \caption{Quality (SC) and Diversity (1-NMI) vs. $\lambda$.}
        \label{fig3a}
	\end{subfigure}
	\quad
	\begin{subfigure}{0.94\columnwidth}
		\centering
		\includegraphics[width=1\columnwidth]{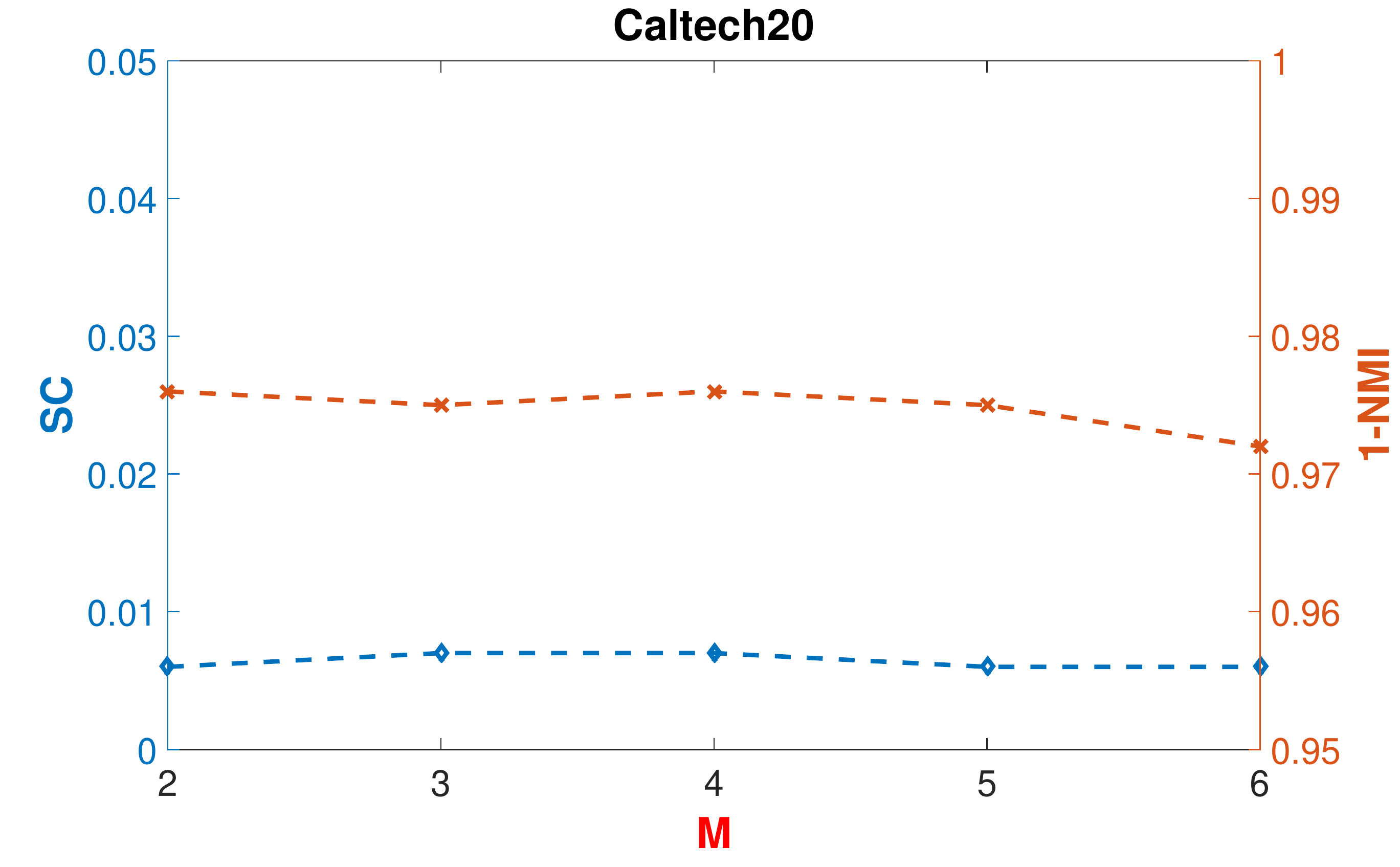}
        \caption{Quality (SC) and Diversity (1-NMI) vs. $M$.}
        \label{fig3b}
	\end{subfigure}
	\label{fig3}
	\caption{Parameter analysis of $\lambda$ and $M$.}
\end{figure*}

\begin{figure*}[t]	
	\centering
	\begin{subfigure}{0.94\columnwidth}
		\centering
    	\includegraphics[width=0.9\columnwidth]{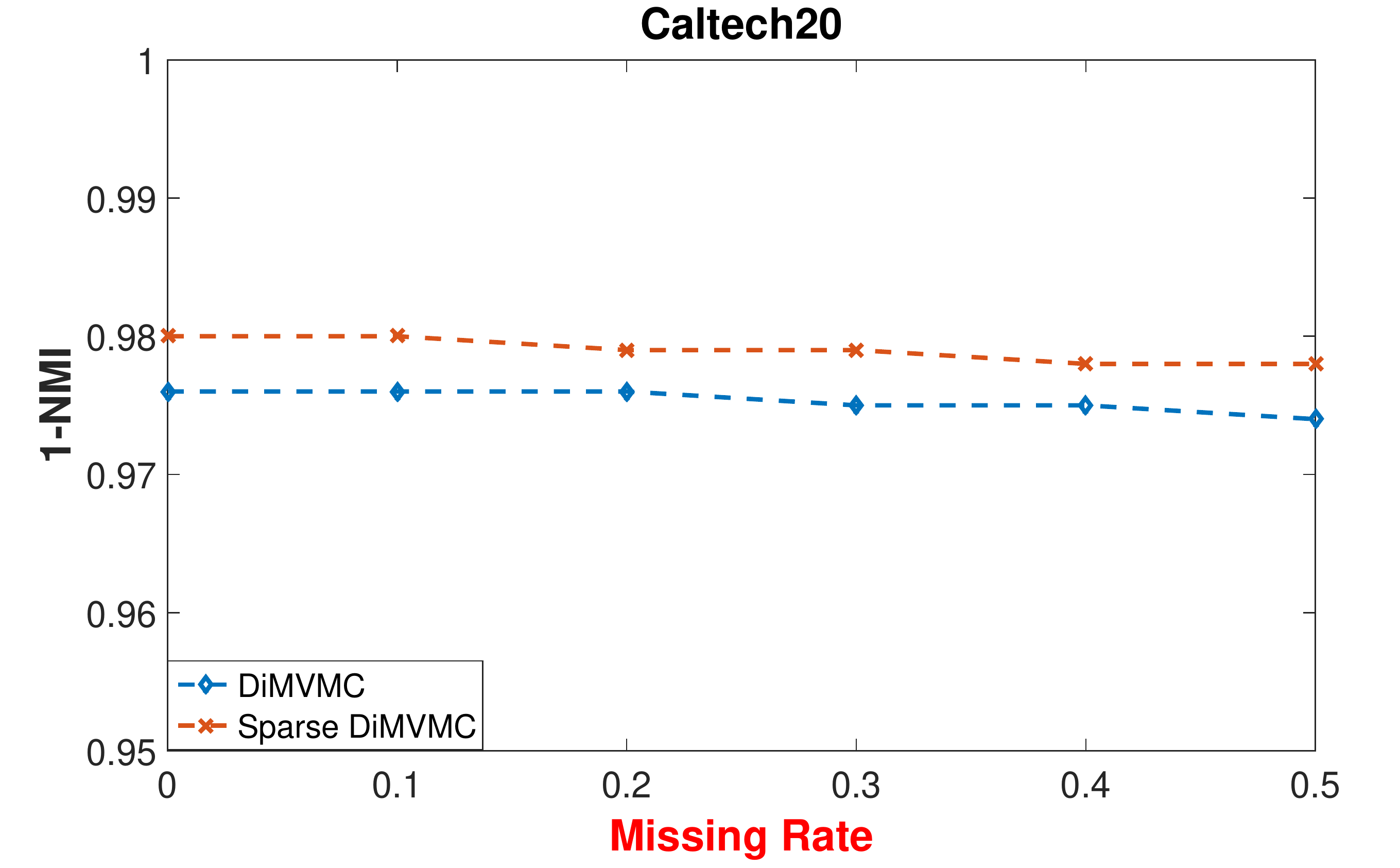}
        \caption{1-NMI vs. missing data rate.}
        \label{fig4a}
	\end{subfigure}
	\quad
	\begin{subfigure}{0.94\columnwidth}
		\centering
		\includegraphics[width=0.9\columnwidth]{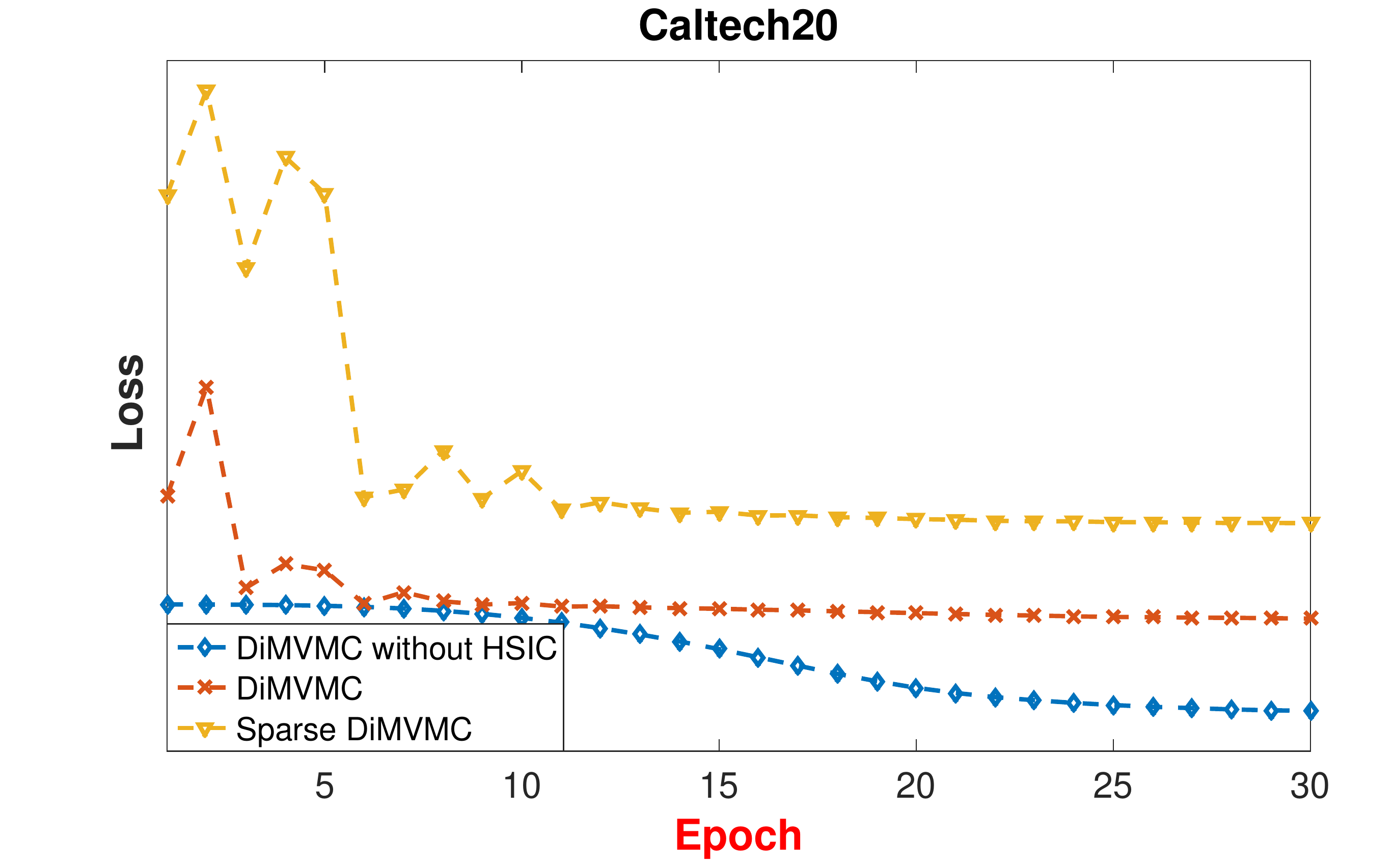}
        \caption{Convergence trend of DiMVMC and its variants.}
        \label{fig4b}
	\end{subfigure}
	\caption{Sparse DiMVMC vs. DiMVMC .}
\end{figure*}

\subsection{Parameter Analysis}
Parameter $\lambda$   balances the generation of multiple subspaces and the diversity control among these subspaces. We study the impact of $\lambda$ by griding it from $10^{-3}$ to $10^{3}$ and plot the variation of quality (SC) and diversity (1-NMI, the larger the better) of DiMVMC on Caltech20 dataset in Figure \ref{fig3a}. We see that: (i) diversity (1-NMI) steadily increases at first and then gradually increases; (ii) the quality (SC) gradually decreases and then keeps relatively stable as $\lambda$ further increases. Overall, SC keeps relatively stable as $\lambda$ varies, but always below the starting point ($\lambda=0$, no diversity control). This pattern is explainable, since the promotion of diversity between clusterings is often associated with the scarification of quality. We can conclude that $\lambda$ indeed helps to boost the diversity between clusterings.

We vary $M$ (number of alternative clusterings) from 2 to 6 on Caltech20 dataset to explore the variation of average quality (SC) and diversity (1-NMI) of multiple clusterings generated by DiMVMC. In Figure \ref{fig3b}, with the increase of $M$, the average quality (SC) fluctuates in a small range while the diversity (1-NMI) decreases slowly. Overall, DiMVMC can obtain $M\geq 2$ alternative clusterings of quality and diversity.

Based on the base DiMVMC, we lead an $l_1$ norm for each common subspace $\mathbf{H}^m$, extend DiMVMC to a sparse DiMVMC. We apply DiMVMC and sparse DiMVMC (with $\alpha=5\times 10^{-2}$) on Caltech20 dataset, and report the results in Figure \ref{fig4a}. Although the SC values of these two models are both around 0.007, sparse DiMVMC has an average NMI (the lower the better) as 0.022, which is nearly 12\% lower than DiMVMC (average NMI 0.025) . This comparison shows that sparsity helps to generate less correlated subspaces (fewer redundant features), and to better control the diversity of alternative clusterings.

\subsection{Convergence and Complexity Analysis}
From Figure \ref{fig4b}, we can find that DiMVMC and Sparse DiMVMC often converge within 15 epochs, while  DiMVMC($\lambda=0$) without diversity control converges in 30 epochs. This trend not only proves the efficiency of our proposed alternative optimization strategy, but also shows that our diversity control term and the added $l_1$ norm do not increase the complexity.

The memory complexity of DiMVMC can be analyzed by two parts. For simplicity, suppose $l$ is the number of layers, $d$ is the dimension for any $\{\mathbf{H}^m\}_{m=1}^M$, and $d_{ave}$ is the average dimension of $V$ views. DiMVMC takes  $\mathcal{O}(NMd+NMVd_{ave})$ to save the data elements, and  $\mathcal{O}(NMVld_{ave})$ to store the network parameters $\{\Theta_m^v\}_{m=1,v=1}^{M,V}$. So the memory complexity of DiMVMC for generating $M$ clusterings on $V$ views is $\mathcal{O}(NM(d+Vd_{ave}+Vld_{ave}))$. Since most multi-view data are typical sparse, the actual space complexity is much smaller.

The time complexity of DiMVMC can be also analyzed by two sub-problems. DiMVMC takes $\mathcal{O}(Nld^{2}d_{ave})$ to update $\Theta_m^v$, and $\mathcal{O}(NVld^{2}d_{ave}+MN^2d)$ to update $\mathbf{H}^m$. So the time complexity of DiMVMC for generating $M$ clusterings on $V$ views is $\mathcal{O}(tMNd(Vldd_{ave}+N))$, where $t$ is the number of iterations. On the other hand, the time complexity of MVMC \cite{yao2019MVMC} is $\mathcal{O}(tMN^{2}V(d+K))$, and that of DMClusts \cite{wei2019multi} is $\mathcal{O}(tMd(MNd_{ave}+MNd+Mdd_{ave}+VNd_{ave}))$. Thus, the time complexity of DiMVMC is quadratic to $N$, due to the use of HSIC term.  MVMC is quadratic to $N$ and DMClusts is linear to $N$. Note, both the memory complexity and time complexity of DiMVMC can drop an order of magnitude via batch optimization technique.

Table \ref{table3} gives the run-time of all compared methods. The compared methods are run on a linux server\footnote{Configuration: Intel Xeon8163, 1TB RAM with NVIDIA Tesla K80.}. All methods are implemented by Matlab2014a, except Nr-kmeans, DMClusts and DiMVMC are implemented by Python. We observe that the three fastest methods are OSC, Dec-kmeans and DMClusts, respectively. OSC and Dec-kmeans do not consider the correlations between views and work on the concatenated views, so they run faster than others. DMClusts employs the efficient semi-NMF \cite{ding2010convex} to decompose multi-view data layer by layer and generates multiple clustering simultaneously. Although MNMF also builds on efficient semi-NMF, it is constrained by the reference clustering when seeking the other clustering, and has a larger run-time than DMClusts. Nr-kmeans needs to update the clustering center many times, so it has a longer run-time than MNMF. MVMC involves with time demanding self-representation learning and  the factorization of multiple representational matrices, so it has a longer run-time than others. Our DiMVMC has the largest run-time, since it has to capture the complex correlations between views and generate multiple clusterings with nonlinear clusters via optimizing multiple decoder networks. However, DiMVMC almost always generate multiple clusterings with  better quality and diversity than these compared methods.

\begin{table*}[t]
\centering
\scriptsize
		\caption{Run-times of compared methods (in seconds).}
\begin{tabular}{l| r r r r r r |r}
	\hline
	&Dec-kmeans  &Nr-kmeans &OSC &MNMF &MVMC &DMClusts &Ours \\
	\hline
    Caltech20 & 334  & 2664 & 180 & 2398 & 1788 &570 & 3296 \\
    Handwriting & 50  & 1212 & 58 & 636 & 2558 &244 & 1106\\
    Reuters  & 1944  & 2096 & 1912 & 10911 & 54095 &4560 & 67583\\
    Mirflickr  & 2386  & 8976 & 263 & 12870 & 53867 &9240 & 78967\\
    \hline
    Total  &4714 &14948 &2413 &16470 &112308 & 14614 &150952\\
    \hline
\end{tabular}
\label{table3}
\end{table*}


\section{Conclusions}
In this paper, we introduced the DiMVMC model to explore alternative clusterings from the ubiquitous incomplete multi-view data. DiMVMC can complete the missing data via a group of decoder networks, and seek multiple shared but diverse subspaces (clusterings therein) by further reducing the overlaps between subspaces. Experimental results on benchmark datasets confirm the effectiveness of DiMVMC. We will explore deep alternative clusterings by merging prior knowledge of different perspectives.

\bibliographystyle{IEEEtran}
\bibliography{DiMVMC_Bib}
\end{document}